\documentclass[letterpaper, 10 pt, journal, twoside]{IEEEtran}
\usepackage[utf8]{inputenc} 
\usepackage[T1]{fontenc}    
\usepackage{url}            
\usepackage{booktabs}       
\usepackage{amsfonts}       
\usepackage{nicefrac}       
\usepackage{microtype}      
\usepackage{xcolor}         

\usepackage{anyfontsize}
\usepackage{amsmath}
\usepackage{amssymb}
\usepackage{wrapfig}
\usepackage{subfig}
\usepackage{caption}
\usepackage{color}

\usepackage{physics}
\usepackage{bm}

\usepackage{algorithm}
\usepackage{algorithmic}
\usepackage{graphicx}

\usepackage{ragged2e}
\usepackage{fp}

\usepackage{ifthen}

\usepackage{pgfplots}
\usepackage{tikz}
\usepackage{tikz-3dplot}

\usepackage{ragged2e}
\usepackage{fp}

\usepackage{ifthen}

\usepackage{tikz}
\usetikzlibrary{
shapes.geometric, 
arrows,
 calc,
 automata,
 positioning,
 decorations,
 decorations.text,
 patterns}
 
 \usetikzlibrary{pgfplots.statistics, pgfplots.colorbrewer} 
\usepackage{pgfplotstable}

\usetikzlibrary{patterns.meta}

\usetikzlibrary{mindmap,snakes}

\usepackage{multirow}

\usepackage{algorithm}
\usepackage{algorithmic}
\usepackage{graphicx}
\usepackage[english]{babel}
\usepackage{array}
\usepackage[export]{adjustbox}

\usepackage{nicefrac}

\usepackage{xcolor}

\usepackage{colortbl}
\usepackage{makecell}
\usepackage{pifont}

\usepackage{mathptmx}

\usepackage[colorlinks=true,
    linkcolor=red,
    linkbordercolor=red,
    urlcolor=mylinkcolor,
    urlbordercolor=white
    ]{hyperref}

\definecolor{mylinkcolor}{rgb}{0.005, 0.3, 0.7}
\definecolor{suppcolor}{rgb}{0.6, 0.0, 0.9}

\usepackage[font=footnotesize]{caption}

\colorlet{baseclr}{gray}
\colorlet{rwclr}{white!90!baseclr}

\colorlet{dotaclr}{black}
\colorlet{dotbclr}{red}

\newcommand{\system}{Drone-Bee}

\newcommand{\cfontpt}[1]{\fontsize{5.7}{\baselineskip}{\selectfont #1}}

\newcommand{\bdota}{\tikz{\node[circle, draw=dotaclr, fill=white!10!dotaclr,scale=0.4]{};}}
\newcommand{\bdotb}{\tikz{\node[circle, draw=dotbclr, fill=white!00!dotbclr,scale=0.4]{};}}

\title{Design, Localization, Perception, and Control for GPS-Denied Autonomous Aerial Grasping and Harvesting \vspace{-0.5ex}}

\author{Ashish~Kumar$^{\dagger}$, Laxmidhar~Behera$^{\dagger}$, \IEEEmembership{Senior Member, IEEE} \vspace{-2.75ex}  \\
\thanks{Manuscript received: December 6, 2023; Accepted February 5, 2024. This paper was recommended for publication by Editor Giuseppe Loianno upon evaluation of the Associate Editor and Reviewers' comments.}
\thanks{$^{\dagger}$EE, Indian Institute of Technology (IIT), Kanpur, India.
{\tt\small \{ashishkumar822@gmail.com,lbehera@iitk.ac.in\}}\\
\textbf{Supplementary:} \textcolor{mylinkcolor} {\footnotesize \url{https://github.com/ashishkumar822/DroneBee}}}
\thanks{Digital Object Identifier (DOI): see top of this page.}
}

\begin{document}

\maketitle

\begin{justify}

\begin{abstract}
%
In this paper, we present a comprehensive UAV system design to perform the highly complex task of off-centered aerial grasping. This task has several interdisciplinary research challenges which need to be addressed at once. The main design challenges are GPS-denied functionality, solely onboard computing, and avoiding off-the-shelf costly positioning systems. While in terms of algorithms, visual perception, localization, control, and grasping are the leading research problems. Hence in this paper, we make interdisciplinary contributions: (\texttt{i}) A detailed description of the fundamental challenges in indoor aerial grasping, (\texttt{ii}) a novel lightweight gripper design, (\texttt{iii}) a complete aerial platform design and in-lab fabrication, and (\texttt{iv}) localization, perception, control, grasping systems, and an end-to-end flight autonomy state-machine. Finally, we demonstrate the resulting aerial grasping system \textit{\system{}} achieving a high grasping rate for a highly challenging agricultural task of apple-like fruit harvesting, indoors in a vertical farming setting (Fig.~\ref{fig:harvesting}). To our knowledge, such a system has not been previously discussed in the literature, and with its capabilities, this system pushes aerial manipulation towards $4^{th}$ generation.
%
%
%
%


\end{abstract}

\end{justify}

\vspace{-0.75ex}
\begin{IEEEkeywords}
Aerial Systems: Applications; Agricultural Automation; Grasping, Dynamic Payloads, Aerial Fruit Harvesting
\end{IEEEkeywords}
\IEEEpeerreviewmaketitle

\vspace{-3.00ex}
\section{Introduction}
\label{sec:intro}

Precision agriculture and vertical plantation are the future of farming. It aims to obtain higher crop yields in a small area of land by performing plantation onto stacked structures, forming tall assemblies (Fig.~\ref{fig:harvesting}). Due to the structured cultivation, we foresee a huge potential of aerial grasping in the harvesting process to increase crop yields and to reduce production costs.
\par
However, aerial grasping is quite difficult \cite{towards}, and is still in its infancy as compared to robotic grasping \cite{semi, acrvsystem2017} due to its convoluted challenges. Some are specific to aerial harvesting, while others are specific to aerial manipulation. For instance, \cite{ollero2021past} extensively reviews aerial manipulators and enlists unsolved challenges. The most crucial ones are \textit{accuracy} of the platform to perform a task and the \textit{decisional autonomy}. Achieving task accuracy is quite essential in the presence of wind gusts or nearby surface perturbations, which are challenging to model and can result in collision \cite{ollero2021past}. Then, for fully autonomous operations, decisional autonomy requires several algorithms to work in conjunction to enable a UAV to make its decision onboard only. However, with the task under consideration, i.e., in-air grasping, the complexity involved is too high, making developing the autonomy engine a difficult process. Finally, cross effects between challenges also exist, e.g. more algorithms require high on-board computational power, which directly affects onboard power, size, and flight time. Since aerial grasping is a subset of aerial manipulation, these unsolved challenges also apply to our case. In this paper, we target them and push the frontiers in this area while targeting aerial grasping. Now, we discuss challenges specific to aerial harvesting and our contributions to advancing the state-of-the-art (SOTA). 
\subsubsection{Task-centric Design}
\label{sec:intro_design}
Hardware design for aerial grasping requires interdisciplinary expertise, e.g. UAV design, sensor system, electronics, etc. Also, the gripper design is task-specific, which is constrained by the power, weight and size of the UAV, turning into an iterative and costly process. 
\par
\textit{Our Contribution:} Although, multi-sensor retractable magnetic gripper \cite{towards}, fixed gripper \cite{nimbrombzirc2020}, multi-finger gripper \cite{lee2021closed}, pipe holding gripper for perching \cite{mclaren2019passive}, multi-DoF arm \cite{ruggiero2015multilayer} exist, works on off-center aerial grasping are lacking in the literature. In this sector, we propose off-center grasping and a gripper design, which are quite novel from an aerial grasping perspective. It is one of the main components of our system, inspired by human-like behaviour for grasping or harvesting tasks.
\input{figures/intro_figure}
\subsubsection{High-Speed and Accurate Visual Perception}
\label{sec:intro_vision}
Visual perception makes aerial grasping feasible in the first place, with deep learning-based detectors \cite{fasterrcnn, ssd} being the modern choice. However, they have complex training and testing phases, with their runtime and accuracy sensitive to postprocessing steps and hyperparameters. Further, detecting fruit-like tiny objects ($ < \sim 15 \times 15$ pixels) from low-resolution CNN features is a challenging research problem. Generally, it is handled via multi-scale detection \cite{fpn} at the cost of increased runtime. However, aerial grasping needs high-speed detection to approach the target and counter the disturbances quickly.
\par
\textit{Our Contribution:} Although post-processing free Transformer-based detectors \cite{detr} are now available, they have large size, slow convergence, and compute and memory-intensive Transformer attention modules which exhaust the low-power devices. In this sector, keeping the challenge of limited onboard computing in mind, we develop a novel single-scale detector free of post-processing and outperforms recent SOTA detectors \cite{yolov8}.
\subsubsection{High-Speed and High-Accuracy Localization}
\label{sec:intro_pos}
Accurate positioning is crucial in UAV control and grasping. For GPS-free operations, using VICON-like motion capture systems or off-the-shelf positioning sensors is not viable, mainly due to their high cost. Also, they lack in building and reusing maps which is a must-have attribute of robotic autonomy. Optical flow sensors are also an option, but they have limited accuracy. Thus, ready-made sensors increase the cost and the number of sensors. Visual markers are unsuitable from a production standpoint due to their installation and algorithmic dependency on them. In contrast, visual SLAM \cite{orb2} offers high metric accuracy, mapping, and customization but is quite compute-intensive. Visual odometry \cite{dmvio} is not viable due to the accumulated drift over time and lack of mapping. 
\par
\textit{Our contributions:}
We address the above challenges and develop a GPU-accelerated stereo visual SLAM design that is the fastest available open-source SLAM system, outperforming state-of-the-art SLAM algorithms.
\subsubsection{Off-centered Dynamic Payload}
\label{sec:intro_control}
It is a critical issue in control systems since off-center grasping requires the gripper to be mounted only along the roll axis, inducing off-center loading. Therefore, as soon as the target is grasped, its weight exerts sudden torque about the pitch axes, resulting in unwanted motion or collision. Although low-weighted centered payload systems exist \cite{pounds2012stability, direct}, control systems for off-centered payload are not explicitly studied, which is of our concern. Moreover, existing controllers are tested in controlled scenarios, e.g. VICON-based positioning, but the evaluation in real scenarios is far more complex.
\par
\textit{Our Contributions:} We develop a new thrust microstepping via acceleration feedback thrust controller that is mass and gravity independent. The thrust controller forms the control system of Drone-Bee and outperforms recent state-of-the-art direct acceleration feedback thrust controller \cite{direct}.
\subsubsection{Precision In-Air Grasping}
\label{sec:intro_inair}
This challenge is listed as a future challenge in \cite{ollero2021past} and is a challenging research problem as it demands reaching and flying near the target with a small tolerance. Also, it relies on vision, control, localization, and depth measurements. Therefore, any error in them can cause grasp failure and collision. There exist many works, such as single arm \cite{heredia2014control}, dual arm \cite{suarez2017anthropomorphic}, parallel manipulator \cite{stephens2022aerial}, human-robot handover task \cite{corsini2022nonlinear}, brick pick-transport-place \cite{towards}, however, none tackles in-air off-center grasping while focusing on design, localization, perception, control and the autonomy engine simultaneously. Although \cite{loianno2018localization} addresses the design, localization, and grasping of magnetic objects, it relies on costly GPS-RTK for localization, and deals with relatively simple scenarios from a vision and control perspective.
\par
\textit{Our Contributions:}
As this problem directly translates to the autonomy engine, we address it via our detection, SLAM, and control systems. We develop an advanced state machine based on our SOTA algorithms, visual servoing and grasping techniques to achieve precision air-grasping.
\subsubsection{Imprecise Depth Measurements}
\label{sec:intro_noisydepth}
The grasping phase heavily relies on depth measurements, acquired from Intel Realsense D$435$i-like devices. Such devices have a tolerance of $\pm 3$cm that is enough to trigger a grasp failure or unwanted grasping of nearby items, thus making it a critical issue.
\par
\textit{Our Contributions:}
Since it is a sensor limitation, we handle it via autonomy engine, i.e. approaching the target slowly when it is near because these devices are relatively accurate at smaller depths. Nonetheless, noise remains in the picture, thus, the probability of losing a target is still not zero.
\subsubsection{Co-existing Subsystems and Only Onboard Computations}
\label{sec:intro_coexisting}
Another challenge is avoiding off-the-shelf positioning systems, only onboard computations, and having GPS-free capability. This requires several co-existing algorithms/sub-systems that mainly include object detection, positioning, control, and grasping. The ones such as object detection and positioning systems, are highly compute-intensive, which prevents concurrent execution of all sub-systems at desired rates without exhausting the onboard computer. Hence, simply deploying existing algorithms is not viable in aerial grasping.
\par
\textit{Our Contributions:}
This challenge poses restrictions on the level of autonomy achievable onboard. We address it via our high-speed detection, SLAM, and control systems. Particularly, these subsystems have been designed in such a way that they can be deployed onboard without computationally taxing the onboard computer. Notably, this is the most critical challenge pointed out by \cite{ollero2021past}. We address this challenge to push aerial manipulation towards $4^{th}$ generation.
\subsubsection{Autonomous Robotic Harvesting}
It is a recent research area to improve crop yields \cite{minneapple, deepfruit, surveying}. Initial efforts are seen in the ground vehicle for crop harvesting by using existing algorithms. However, developing such systems is very difficult due to the complex nature of the harvesting tasks, varying from crop to crop. 
This also applies to UAV-based harvesting, which is even more convoluted due to the previously discussed challenges and complex onboard flight autonomy. This is one of the primary reasons why an end-to-end UAV-based harvesting system is still not visible in the literature.
\par
\textit{Our Contributions:} Hence, in this paper, we have developed a complete hardware and algorithmic solution by pushing the state-of-the-art in multiple domains, called \textit{Drone-Bee}. We demonstrate it performing the complicated task of aerial grasping indoors and outdoors. To our knowledge, this is the first open-source aerial harvesting system in the literature.
\subsubsection{Lack of Realistic Experimental Setup for Harvesting}
\label{sec:intro_setup}
Despite many robotic harvesting solutions, none addresses the issue of experimental setup. Since developing a robotic harvesting system is itself a challenging research area, it is not possible to go into the fields or farms every time to conduct experiments. This brings up the challenge of accelerating system development without going into the fields and round-the-clock experimentation.
\par
\textit{Our contributions:} Hence, we develop an easy-to-build in-lab harvesting setup for researchers to perform experiments round-the-clock, accelerating the system development from multiple fronts, e.g. hardware and software, autonomy, etc.

\section{Hardware Design}
\label{sec:sysdesign}
In this section, we address the hardware design challenges of aerial grasping and describe our contributions. Our motivations are low-cost hardware, ease of component accessibility, in-lab fabrication, and rapid prototyping to accelerate research.
\subsection{UAV Platform}
\label{sec:uavdes}
This is the major source of cost which can be lowered by developing a customized platform. Although our solution is UAV-agnostic, in this work, we use DJI Matrice-$100$ quadrotor (Table~\ref{tab:specs}) while focusing more on the aerial grasping motive. The system can carry a payload of $1.2$Kg$@70\%$ motor spin. More weight is avoided to prevent motor heating. We control the UAV via roll, pitch, yaw, and thrust commands, which are generated by our control system (Sec.~\ref{sec:control}). We use DJI Onboard-SDK (OSDK) to send these commands to the flight controller and to receive IMU data from it.
\input{figures/uav}

\begin{table}[!t]
\centering
\caption{UAV platform specifications.}
\label{tab:specs}
\arrayrulecolor{white!70!black}
\scriptsize
\setlength{\tabcolsep}{1.0pt}
\vspace{-0.75ex}

\begin{tabular}{c c c c}
\hline
\multicolumn{1}{c}{Attribute} & \multicolumn{1}{|c}{Specification} & \multicolumn{1}{||c}{Attribute} & 
\multicolumn{1}{|c}{Specification} \\ \cline{1-4}
\multicolumn{1}{c}{\cfontpt{Size}} & \multicolumn{1}{|c}{\cfontpt{$0.90$m$\times$ $0.90$m$\times$ $0.45$m}} & \multicolumn{1}{||c}{\cfontpt{Operating Voltage}} &  \multicolumn{1}{|c}{\cfontpt{$22$V}} \\
\multicolumn{1}{c}{\cfontpt{Size w/ Gripper}} & \multicolumn{1}{|c}{\cfontpt{$1.5$m $\times$ $0.90$m $\times$ $0.45$m}} & \multicolumn{1}{||c}{\cfontpt{Hover time}} & \multicolumn{1}{|c}{\cfontpt{$20$ minutes}} \\
\multicolumn{1}{c}{\cfontpt{Weight}} & \multicolumn{1}{|c}{\cfontpt{$2.5$Kg}} & \multicolumn{1}{||c}{\cfontpt{Hover time w/ Gripper}} & \multicolumn{1}{|c}{\cfontpt{$15$ minutes}} \\
\multicolumn{1}{c}{\cfontpt{Weight w/ Gripper}} & \multicolumn{1}{|c}{\cfontpt{$3.4$Kg}} & \multicolumn{1}{||c}{\cfontpt{Onboard Computer}} & \multicolumn{1}{|c}{NVIDIA Jetson-NX} \\
\multicolumn{1}{c}{\cfontpt{Rotor diameter}} & \multicolumn{1}{|c}{\cfontpt{$0.46$m}} & \multicolumn{1}{||c}{\cfontpt{Communication}} & \multicolumn{1}{|c}{\cfontpt{UART $@921600$ Baud}} \\
\multicolumn{1}{c}{\cfontpt{Rotor-to-Rotor distance}} & \multicolumn{1}{|c}{\cfontpt{$0.35$m}} & \multicolumn{1}{||c}{\cfontpt{Stereo-Rig}} & \multicolumn{1}{|c}{\cfontpt{$2 \times$~$@432 \times 240, 30$Hz}} \\
 \hline
\end{tabular}
\vspace{-3ex}

\end{table}
\subsection{Gripper Design}
\label{sec:gripperdes}
Our gripper design is primarily inspired by two reasons; \textit{First,} fruits in orchards or trees lie vertically, thus approaching them from the top and performing centered grasping \cite{loianno2018localization, lee2021closed} is not feasible. \textit{Second}, humans perform grasping by extending their multi-DoF arm horizontally in similar cases. However, the multi-DoF arm in aerial platforms has considerable stability issues \cite{ruggiero2015multilayer}. This gives rise to our novel idea of off-center grasping to approach the fruits horizontally.
\par
To this end, we propose a lightweight and compact, fixed-length,  off-center gripper having a human-like three-finger hand jaw with a precisely controllable opening (Fig.~\ref{fig:gripper}). Our design reduces gripper complexity and platform instability while increasing the target reachability. In terms of lifting capacity, it can easily hold apple-like spherical items having a radius in the range $\sim 3-6$cm, and weight up to $2.5$ kg.
\par
The gripper mainly has four parts: \textit{fingers}, \textit{wrist}, \textit{arm}, and \textit{actuator} (Fig.~\ref{fig:cads}). Each finger consists of two hinges, one attached to the wrist and the other attached to a \textit{finger coupler} (Fig.~\ref{fig:cadfingercoupler}). A push action on the coupler produces outward torque on the fingers, leading to the gripper opening, while a pull action produces inward torque, leading to the gripper closure. The push and pull actions are realized by linking the coupler with a carbon-fiber tube of outer diameter $6$mm, called \textit{coupler link}. The coupler link runs inside another carbon-fiber tube of outer diameter $16$mm, called \textit{arm}. At one end of the arm, the wrist is mounted, whereas another end is attached to a \textit{plate}, on which the actuator is mounted. The actuator provides linear motion to the coupler link that precisely controls the gripper opening or closure. The actuator and the coupler link are coupled via a $3$D printed part (Fig.~\ref{fig:gripper}).
\input{figures/gripper}
\input{figures/cad}
\par
The actuator comes from Actuonix P$16$-P with $10$cm stroke length and voltage feedback to obtain the current extruded length. The control circuit is developed in-lab, consisting of a PIC$18$F$2550$ microcontroller used for ADC conversion of the feedback signal and to drive the actuator. The microcontroller communicates with the onboard computer via a serial link.
\par
We mount the gripper at the bottom of the UAV. The arm of the gripper is quite long ($84$cm from the UAV center), which introduces severe vibrations because it is left open at the wrist end. To suppress them, we design a gripper support structure (Fig~\ref{fig:cadgrippersupport}) which holds the arm firmly (Fig.~\ref{fig:cadgripper},~\ref{fig:cadgripperwithsupport}). Please see the video for a visual illustration of the vibrations.
%
\par
Our gripper fulfils the size, weight and power constraints, and has fairly easy assembling, thanks to its modular design.
\subsection{Vision Sensors}
\label{sec:sensors}
%
%
%
Our positioning system is based on Stereo visual SLAM; Hence, from a cost and customization standpoint, we take two Lenovo $300$ FHD cameras and extract their PCB. Then we mount the PCBs onto a $3$D-printed part to form a stereo-rig (Fig.~\ref{fig:cadstereo},~\ref{fig:realstereo}). We use OpenCV \cite{opencv} for calibrating the rig.
\subsection{Communication Device and Emergency Kill-Switch}
We use ASUS GT-AX-$11000$ wireless router to link the ground station and the UAV for real-time data logging and monitoring. We develop a \textit{software kill-switch} that is available on the ground station and halts the UAV in case of fatality.
\subsection{Onboard Computing Infrastructure}
We use one NVIDIA Jetson Xavier NX, a low-powered $10|20$W embedded computer with $6$ CPU cores, $384$ GPU cores for parallel computing, $8$GB RAM and a dual-band WiFi, and four USB $3.0$ ports for interfacing peripherals such as USB-Serial link, Intel Realsense D$435$i depth sensor. The flight controller communicates with the Jetson NX via a UART available at \texttt{/dev/ttyTHS0} in the operating system. 
\subsection{Harvesting Region and Target Fruit}
\label{sec:setup}
Harvesting farms may be unavailable near the research lab (Sec.~\ref{sec:intro_setup}). We address this challenge and propose a harvesting setup resembling plants in vertical farming or orchards. We use a wooden trellis that is attached to the harvesting setup skeleton and is wrapped with artificial leaves to simulate plants. The setup stands tall at $1.50$m, while the harvesting region begins at $0.80$m and measures $2.20$m $\times 0.70$m (Fig.~\ref{fig:harvesting}).
\par
Our harvesting setup offers several benefits: \textit{First}, it can be kept indoors, allowing the testing of the algorithms and dataset collection. \textit{Second}, it allows the iterative system design process (Sec.~\ref{sec:intro_design}), i.e. build-modify-test without going into the fields and waiting for favourable weather conditions. These attributes speed up the overall system design and testing.
\par
We use apple fruit as the target object, weighing $300$gms and having radii $\sim 6$cm. In this work, we keep only one type of target to avoid too much complexity in the system design, because the UAV-based harvesting is already quite complex.
%

\section{Flight Autonomy Sub-Systems}
\label{sec:method}
Flight Autonomy is the most fundamental and state-of-the-art challenge to realize $4^{th}$ generation of aerial manipulators \cite{ollero2021past}. In this work, we propose an advanced flight autonomy engine based on our high-speed algorithms, which outperforms state-of-the-art algorithms while catering for the constraint of limited onboard computing power. Below, we describe our algorithm which we develop for the system autonomy.
%
%
\subsection{High-Speed Accurate Localization}
\label{sec:localization}
\par
A $4^{th}$ generation system would require localization without visual markers or costly off-the-shelf GPS-RTK sensors \cite{loianno2018localization}, leading to challenging research problems since speed, metric accuracy, and precision are critical for control and grasping (Sec.~\ref{sec:intro_pos}) while having high speed. Hence, we base our localization module on stereo visual SLAM due to its high metric accuracy, however, its high computing costs \cite{orb2} becomes a bottleneck for Jetson NX-like devices in the presence of other algorithms. 
\par
Hence, we develop a GPU-accelerated, resource-efficient and accurate stereo visual SLAM system (Fig.~\ref{fig:slampipeline}, \cite{jetsonslam}) reaching beyond $60$Hz $@432\times240$ even at eight scales on Jetson NX alongside other algorithms. It is based on our novel components of (\textit{i}) Bounded Rectification to prevent tagging of non-corners as corners, and (\textit{ii}) Pyramidal Culling and Aggregation (\texttt{PyCA}) to obtain robust features at high speeds by harnessing a GPU device. PyCA is based on our novel techniques of feature culling, pyramidal feature aggregation, efficient GPU warp allocation via multi-location per thread culling and thread-efficient warp allocation. (\textit{iii}) We also develop synchronized shared memory that turns our SLAM system resource-efficient. Our SLAM system is the fastest available accurate and GPU-accelerated system (Sec.~\ref{sec:evalalgo}).
\par
Further, to obtain a high-frequency state estimation of the UAV, we use Extended Kalman Filter (EKF) and fuse the onboard Inertial-Measurement-Unit (IMU) data with $6$-DoF state ($x,y,z,\phi,\theta,\psi$) provided by the SLAM system.
\input{figures/slam_arch}
\subsection{Modelling and Control System}
\label{sec:control}
\par
Our UAV platform is shown in Fig.~\ref{fig:m100}. We define a \textit{world frame} $\mathcal{F}_W =[\bm x_W, \bm y_W, \bm z_W]$ fixed at the take-off point, and a UAV \textit{body frame} $\mathcal{F}_B=[\bm x_B, \bm y_B, \bm z_B]$ coinciding with the UAV's Center-of-Gravity (CoG). $\mathcal{F}_B$ is described relative to $\mathcal{F}_W$ via a position vector $\bm p_B=[x, y, z]^T \in \mathbb{R}^3$ and a rotation matrix $\bm R_B \in SO3$. Both $\mathcal{F}_W$ and $\mathcal{F}_B$ follows \textit{\textcolor{red}{Front}-\textcolor{green}{Left}-\textcolor{blue}{Up}} convention, denoting their $\{\bm x, \bm y, \bm z \}$ axes, respectively. The UAV's velocity, acceleration, and angular velocity are denoted as $\bm v_B, \bm a_B, \bm \omega_B \in \mathbb{R}^3$, expressed in $\mathcal{F}_W$, except $\omega_B$ is in $\mathcal{F}_B$.
\par
The UAV motion is governed by the \textit{thrust} $ f_B \in \mathbb{R}$ applied parallel to $\bm z_B$, and \textit{torques} $\bm \tau_B \in \mathbb{R}^3$ operating in $\mathcal{F}_B$. These are controlled via four commands: namely \textit{thrust, roll, pitch, yaw} or simply $\{f_B, \phi, \theta, \psi\}$. The above model description leads to the following rigid body dynamics of the UAV:
\begin{equation}
\footnotesize
\bm{\dot{p}}_B = \bm{v}_B,~~~\bm{\dot{v}}_B = \bm{a}_B \label{eq:dynamicsvel} \\
\end{equation}
\begin{equation}
\footnotesize
\bm{f}_B = m \bm{a}_B + mg - \bm{f_e} \label{eq:mi}
\end{equation}
\begin{equation}
\footnotesize
\bm{\dot{R}}_{B} = \bm S (\bm \omega_B) \bm{R}_{B} \label{eq:dynamicsangvel},~~~\bm J_B \bm{\dot{\omega}}_B = -\bm S (\bm \omega_B) \bm J_R \bm{\omega}_B + \bm \tau_BS
\end{equation}
where, $\bm f_e \in \mathbb{R}^{3}$ is external disturbance, $m$ is the UAV mass, and $J_B$ is the inertia matrix. $\bm S(\bm \cdot) $ is a skew-symmetric matrix. A `$\star$' superscript denotes the desired value of any variable.
\par
The thrust controller in the above model (Eq.~\ref{eq:mi}) poses a challenge in aerial grasping due to the dynamic payloads, wind drafts, and battery issues (Sec.~\ref{sec:intro_control}). We address them simultaneously via our novel Thrust Microstepping via Accelerometer Feedback  \cite{tmdc} technique. It accurately estimates the thrust in the position controller of the quadrotor control, even without the knowledge of UAV mass and gravity, outperforming the existing thrust controllers (Eq.~\ref{eq:mi}) \cite{direct}. In that case, the definition of our thrust controller becomes:
\begin{align}
\bm f^\star_B &= \alpha \bm e_{\bm a} + \beta \dot{\bm e}_{\bm a} + \bm f^{\star}_{B_{t-1}}
\end{align}
where $\bm \alpha, \bm \beta$ are the tunable parameters, $ \bm e_{\bm a} = \bm a^\star_B - \bm a_B$, and $\bm f^\star_B$ is the desired thrust vector to attain $\bm a^\star_B$. See \cite{tmdc} for details.
\par
Further, aerial grasping necessitates a velocity controller for visual servoing tasks. Hence, we developed a position controller based on the proposed thrust controller. Based on the desired position or velocity, the desired acceleration $\bm a^\star_B$ is obtained that is fed to the thrust controller. The obtained thrust is fed to \cite{geometrictracking} to calculate desired orientation. Thus, the overall control system outputs $\{f_B^\star, \phi^\star, \theta^\star, \psi^\star\}$ which is sent to the attitude controller via OSDK (Sec.~\ref{sec:uavdes}). Our control system offers independent control in all three axes, i.e. an axis can operate in position or velocity mode irrespective of the others, facilitating precise navigation and visual servoing. This is the major novelty of our control system design, along with its mass and gravity-agnostic nature. See Sec.~\ref{sec:evalalgo} for the comparison with the state-of-the-art.
\input{figures/ffd_figure}
\subsection{High-Speed Object Detection}
\label{sec:detector}
Object detection is a critical component in aerial grasping because any error in it can cause incorrect target grasp or even failure. Hence, we use deep learning-based detectors for their accuracy, but they suffer from many limitations in our context (Sec~\ref{sec:intro_vision}); thus, we redesign the detection head in CNN-based detectors because it consists of most of the hyperparameters and post-processing steps. Our detector FFD \cite{ffd} is exceptionally lightweight, single-stage, single-scale, free of anchor-box and NMS while having much-simplified training and testing phases (Fig~\ref{fig:objdetect}). Thus, FFD reaches $100$FPS$@$FP$32$-precision on Jetson NX while co-existing with the other time-critical sub-systems, such as localization, control, and grasping.
\par
In FFD, an image is forwarded through a CNN backbone with a progressive reduction in the spatial size. The backbone is customized VGG \cite{vgg}, having five stages with $\{2,2,3,3,4\}$ layers and $\{16,32,64,128,256\}$ neurons per stage, operating at a stride of two. The backbone output is fed to our novel Latent Object Representation (\texttt{LOR}) module, which aggregates global information via Cross Channel Global Context (\texttt{CCGC}), and outputs queries similar to Transformer-based detectors but without using Transformers. Each query represents an object which refined using Query Transformation (\texttt{QT}). The queries are finally passed through two Feed Forward Neural-Networks (FFN), producing classification scores and bounding boxes.
\par
Further, we adapt synthetic scene synthesis to generate vast training data.  This is a scalable approach in real time as collecting and labelling fruit images is a time-consuming task since these images consist of many instances. 
\par
Our FFD outperforms many state-of-the-art multi-scale detectors in various aspects, i.e. speed and accuracy. See Sec.~\ref{sec:evalalgo} for the evaluation.

\section{State-Machine for Aerial Grasping Autonomy}
\label{sec:sys}
\begin{figure}[!t]
\centering
\colorlet{posdclr}{white!0!green}
\colorlet{veldclr}{white!0!cyan}
\colorlet{accjdclr}{white!0!magenta}

\colorlet{slamdclr}{white!0!yellow}
\colorlet{slamclr}{white!90!black}

\colorlet{ekfdclr}{white!0!blue}
\colorlet{ekfclr}{white!90!black}

\colorlet{adpddclr}{orange!90!white}
\colorlet{adpdclr}{white!90!black}

\colorlet{attdclr}{white!90!black}
\colorlet{attclr}{white!90!black}

\colorlet{mxrdclr}{white!90!black}
\colorlet{mxrclr}{white!90!black}

\colorlet{uavdclr}{white!90!black}
\colorlet{uavclr}{white!90!black}

\colorlet{drawclr}{white!40!black}

\colorlet{hwdclr}{cyan!90!white}
\colorlet{hwclr}{white!90!black}

\colorlet{ftfddclr}{magenta!50!black}
\colorlet{ftfdclr}{white!90!black}

\colorlet{griprdclr}{red!80!white}
\colorlet{griprclr}{white!90!black}

\colorlet{statedclr}{green!40!black}
\colorlet{stateclr}{white!90!black}

\begin{tikzpicture}

\node [scale = 0.75]
{
\tikz{
\node (outer) [draw=white!50!black, rounded corners=0.3ex, minimum width=71ex, minimum height=24ex, xshift=0ex, yshift=0ex]{};
\node (uav) [draw=uavdclr, rounded corners=0.3ex, xshift=25ex, yshift=0ex]{\includegraphics[scale=0.030]{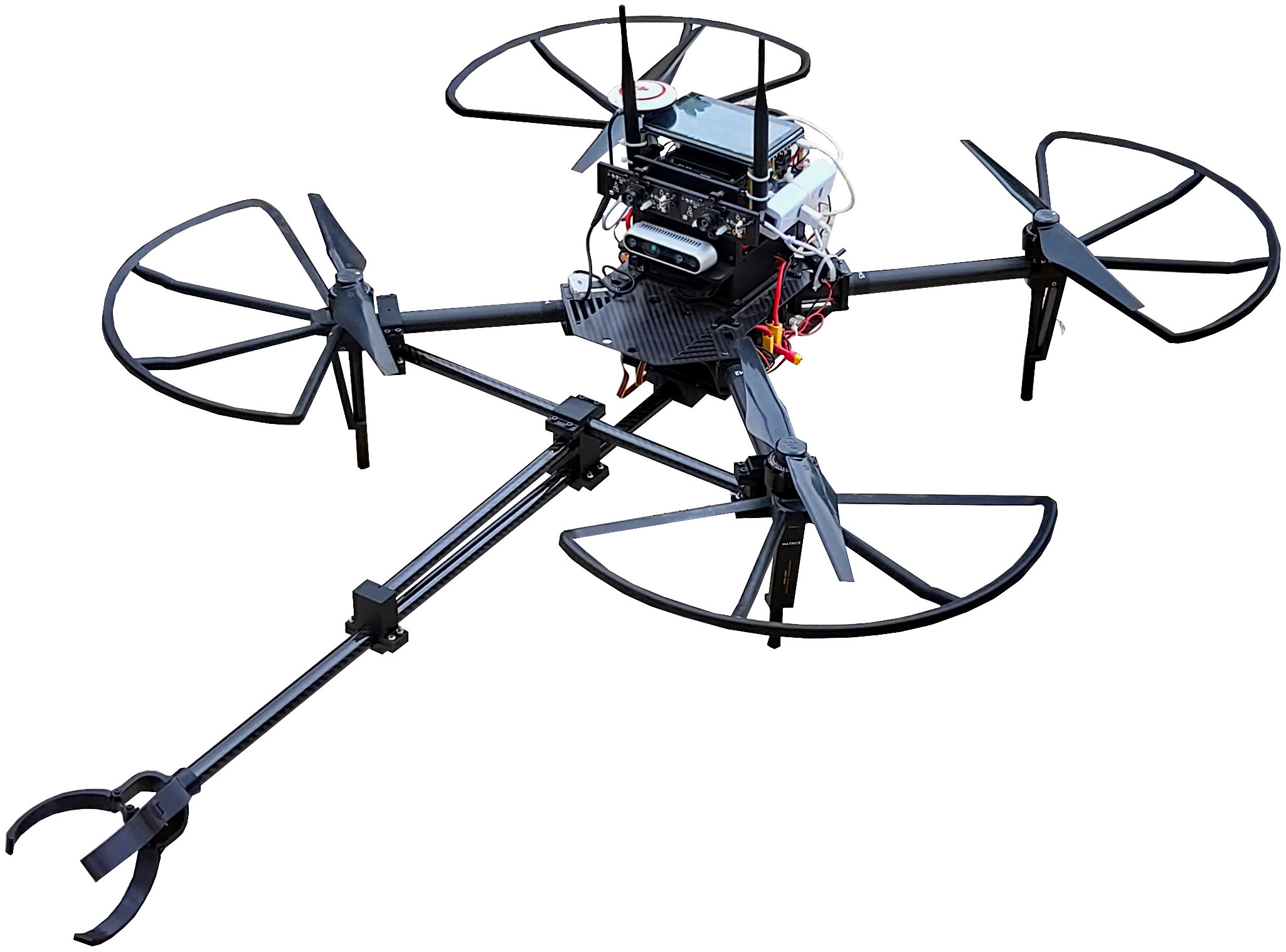}};
\node (vslam) [draw=slamdclr, fill= slamclr, rounded corners=0.3ex, xshift=-27ex, yshift=-9ex, minimum height=3.5ex]{SLAM};
\node (ekf) [draw=ekfdclr, fill= ekfclr, rounded corners=0.3ex, xshift=-2ex, yshift=-9ex, minimum height=3.5ex]{State Mixer (EKF)};
\node (adpd) [draw=adpddclr, fill= adpdclr, rounded corners=0.3ex, minimum height=3.5ex, xshift=25ex, yshift=-9ex]{\shortstack{UAV Control}};
\node (rig) [draw=hwdclr, fill= hwclr, rounded corners=0.3ex, minimum height=3.5ex, xshift=-28ex, yshift=9ex]{\shortstack{Stareo Rig}};
\node (imu) [draw=hwdclr, fill= hwclr, rounded corners=0.3ex, minimum height=3.5ex, xshift=-18ex, yshift=-3ex]{\shortstack{IMU}};
\node (realsense) [draw=hwdclr, fill= hwclr, rounded corners=0.3ex, minimum height=3.5ex, xshift=-10ex, yshift=9ex]{\shortstack{Intel Realsense}};
\node (ftfd) [draw=ftfddclr, fill= ftfdclr, rounded corners=0.3ex, minimum height=3.5ex, xshift=6ex, yshift=9ex]{\shortstack{Detector}};
\node (state) [draw=statedclr, fill= stateclr, rounded corners=0.3ex, minimum height=6ex, xshift=-0ex, yshift=0ex]{\shortstack{Fruit Harvesting \\ State machine}};
\node (gripper) [draw=griprdclr, fill= griprclr, rounded corners=0.3ex, minimum height=3.5ex, xshift=25ex, yshift=9ex]{\shortstack{Gripper Control}};
\draw [thick, ->] (rig) -- ($(vslam.north)-(1ex,0ex)$);
\draw [thick, ->] (vslam) -- (ekf);
\draw [thick, ->] (imu) -- ($(imu.east)+(3ex,0ex)$)  -- ($(imu.east)+(3ex,-2.5ex)$) -| ($(ekf.north)-(4ex,0ex)$);
\draw [thick, ->] ($(ekf.north)+(2ex,0ex)$) -- (state);
\draw [thick, ->] (realsense) -- (ftfd);
\draw [thick, <->] (state) -- ($(state.north)+(0ex,2.5ex)$)  -| (ftfd);
\draw [thick, <->] (state) -- ($(state.east)+(3.5ex,0ex)$)  |- (gripper);
\draw [thick, ->] (ekf) -- (adpd);
\draw [thick, ->] (adpd) -- (uav);
\draw [thick, <->] ($(state.south)+(5.5ex,0ex)$) -- ($(state.south)+(5.5ex,-3.25ex)$) -| ($(adpd.north)-(2.0ex,0ex)$) ;
\draw [thick, ->] (gripper) -- (uav);
}
};
\end{tikzpicture}
\newcommand{\hwa}{\raisebox{0.1ex}{\tikz{\node(a)[line width=0.2ex,draw=uavdclr, scale=0.75]{};}}}
\newcommand{\hwb}{\raisebox{0.1ex}{\tikz{\node(a)[line width=0.2ex,draw=hwdclr, scale=0.75]{};}}}
\newcommand{\swa}{\raisebox{0.1ex}{\tikz{\node(a)[line width=0.2ex,draw=slamdclr, scale=0.75]{};}}}
\newcommand{\swb}{\raisebox{0.1ex}{\tikz{\node(a)[line width=0.2ex,draw=ekfdclr, scale=0.75]{};}}}
\newcommand{\swc}{\raisebox{0.1ex}{\tikz{\node(a)[line width=0.2ex,draw=adpddclr, scale=0.75]{};}}}
\newcommand{\swd}{\raisebox{0.1ex}{\tikz{\node(a)[line width=0.2ex,draw=ftfddclr, scale=0.75]{};}}}
\newcommand{\swe}{\raisebox{0.1ex}{\tikz{\node(a)[line width=0.2ex,draw=griprdclr, scale=0.75]{};}}}
\newcommand{\swf}{\raisebox{0.1ex}{\tikz{\node(a)[line width=0.2ex,draw=statedclr, scale=0.75]{};}}}
\vspace{-2.5ex}
\caption{Connectivity between different harware (\protect\hwa, \protect\hwb) and software (\protect\swa, \protect\swb, \protect\swc, \protect\swd, \protect\swe, \protect\swf) components.}
\label{fig:infoflow}
\vspace{-1ex}
\end{figure}

%
%
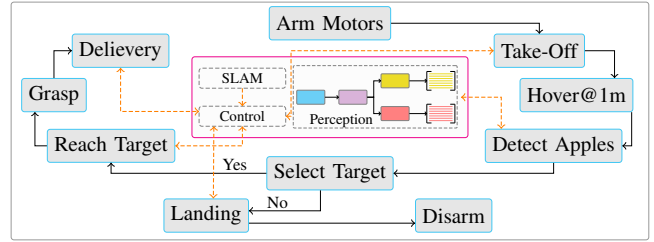
\begin{figure}[!t]
\centering
\begin{tikzpicture}


\colorlet{drawclr}{white!40!cyan}
\colorlet{fillclr}{white!90!black}
\colorlet{middrawclr}{white!0!magenta}
\colorlet{midfillclr}{white!98!gray}

\node [scale = 0.8]
{
\tikz{
\node (outer_box) [draw=white!60!black,rounded corners=0.3ex,minimum width=67ex, minimum height = 25ex, xshift = 0ex,yshift=-4.5ex,]{};
%
\node (arm) [draw=drawclr, fill=fillclr, rounded corners=0.3ex,minimum width=1.5ex, minimum height = 3.5ex, xshift = 0ex, yshift=26] {Arm Motors};
\node (takeoff) [draw=drawclr, fill=fillclr, rounded corners=0.3ex,minimum width=1.5ex, minimum height = 3.5ex, xshift = 22ex, yshift=13] {Take-Off};
\node (hover) [draw=drawclr, fill=fillclr, rounded corners=0.3ex,minimum width=1.5ex, minimum height = 3.5ex, xshift = 25.9ex, yshift=-8] {Hover@$1$m};
\node (detect) [draw=drawclr, fill=fillclr, rounded corners=0.3ex,minimum width=1.5ex, minimum height = 3.5ex, xshift = 23.5ex, yshift=-32] {Detect Apples};
\node (target) [draw=drawclr, fill=fillclr, rounded corners=0.3ex,minimum width=1.5ex, minimum height = 3.5ex, xshift = 0ex, yshift=-45] {Select Target};
\node (vs) [draw=drawclr, fill=fillclr, rounded corners=0.3ex,minimum width=1.5ex, minimum height = 1ex, xshift = -23ex, yshift=-32] {Reach Target};
\node (grasp) [draw=drawclr, fill=fillclr, rounded corners=0.3ex,minimum width=1.5ex, minimum height = 3.5ex, xshift = -29.0ex, yshift=-8] {Grasp};
\node (delivery) [draw=drawclr, fill=fillclr, rounded corners=0.3ex,minimum width=1.5ex, minimum height = 3.5ex, xshift = -22ex, yshift=13] {Delievery};
\node (land) [draw=drawclr, fill=fillclr, rounded corners=0.3ex,minimum width=1.5ex, minimum height = 3.5ex, xshift = -13ex, yshift=-65] {Landing};
\node (disarm) [draw=drawclr, fill=fillclr, rounded corners=0.3ex,minimum width=1.5ex, minimum height = 3.5ex, xshift =13ex, yshift=-65] {Disarm};
\node (mid) [draw=middrawclr,fill=midfillclr, rounded corners=0.3ex,minimum width=29ex, minimum height = 8.5ex, xshift = 0ex, yshift=-9] {};
\node (nwarch) [yshift = 1.0ex,xshift=5ex,yshift=-3ex, scale = 0.82]
{\tikz{
\node (outer_box) [dashed, line width= 0.1pt,dash pattern=on 2pt off 1pt, draw=white!50!black,rounded corners=0.25mm,minimum width=21.0ex, minimum height = 8ex, xshift = -4ex,yshift=0.05ex,]{};
%
\node (nwarch) [yshift = 0ex,xshift=-3.7ex,scale=0.6]
{\tikz{
\colorlet{tensorclr}{white!90!blue}
\colorlet{tensordclr}{white!50!black}
\colorlet{maxclr}{white!80!black}
\colorlet{convclr}{white!80!green}
\colorlet{reluclr}{white!90!black}
\colorlet{eltclr}{white!80!blue}
\colorlet{softmaxclr}{white!80!blue}
\colorlet{drawclr}{white!99!black}
\colorlet{posencclr}{white!50!orange}
\colorlet{transposeclr}{white!60!red}
\colorlet{ffncclr}{black!10!yellow}
\colorlet{ffnbclr}{white!50!red}
\node (backbone) [draw=gray, fill=white!50!cyan, rounded corners=0.25ex, minimum width=6ex, minimum height=3ex, xshift = 0ex, scale=1.0]{};
\node (lor) [draw=gray, fill=white!70!violet, rounded corners=0.25ex, minimum width=6ex, minimum height=3ex, xshift = 9ex, scale=1.0]{};
\node (ffnc) [draw=gray, fill=ffncclr, rounded corners=0.25ex, minimum width=6ex, minimum height=3ex, xshift = 18ex, yshift=3.5ex, scale=1.0]{};
\node (ffnb) [draw=gray, fill=ffnbclr, rounded corners=0.25ex, minimum width=6ex, minimum height=3ex, xshift = 18ex, yshift=-3.5ex,  scale=1.0]{};
\node (mtx1) [fill=none, xshift=27.0ex, yshift=3.5ex, scale=1.0]
{
\FPeval{sx}{6}
\FPeval{sy}{4}
\tikz{
\draw [] (0ex,0ex) -- (1ex, 0ex);
\draw [] ($(\sx ex, 0ex)-(1ex,0ex)$) -- (\sx ex, 0ex);
\draw [] (0ex,0ex) -- (0ex, -\sy ex);
\draw [] (\sx ex,0ex) -- (\sx ex, -\sy ex);
\draw [] (0ex, -\sy ex) -- (1ex, -\sy ex);
\draw [] ($(\sx ex, -\sy ex)-(1ex, 0 ex)$) -- (\sx ex, -\sy ex);
\draw [black!10!yellow,  thick] (0.5ex, -0.5ex) -- ($(\sx ex, -0.5ex)-(0.5ex,0ex)$);
\draw [black!10!yellow,  thick] (0.5ex, -1.0ex) -- ($(\sx ex, -1.0ex)-(0.5ex,0ex)$);
\draw [black!10!yellow,  thick] (0.5ex, -1.5ex) -- ($(\sx ex, -1.5ex)-(0.5ex,0ex)$);
\draw [black!10!yellow,  thick] (0.5ex, -2.0ex) -- ($(\sx ex, -2.0ex)-(0.5ex,0ex)$);
\draw [black!10!yellow,  thick] (0.5ex, -2.5ex) -- ($(\sx ex, -2.5ex)-(0.5ex,0ex)$);
\draw [black!10!yellow,  thick] (0.5ex, -3.0ex) -- ($(\sx ex, -3.0ex)-(0.5ex,0ex)$);
\draw [black!10!yellow,  thick] (0.5ex, -3.5ex) -- ($(\sx ex, -3.5ex)-(0.5ex,0ex)$);
%
%
}
};
\node (mtx2) [fill=none, xshift=27.0ex, yshift=-3.5ex, scale=1.0]
{
\FPeval{sx}{6}
\FPeval{sy}{4}
\tikz{
\draw [] (0ex,0ex) -- (1ex, 0ex);
\draw [] ($(\sx ex, 0ex)-(1ex,0ex)$) -- (\sx ex, 0ex);
\draw [] (0ex,0ex) -- (0ex, -\sy ex);
\draw [] (\sx ex,0ex) -- (\sx ex, -\sy ex);
\draw [] (0ex, -\sy ex) -- (1ex, -\sy ex);
\draw [] ($(\sx ex, -\sy ex)-(1ex, 0 ex)$) -- (\sx ex, -\sy ex);
\draw [ffnbclr,  thick] (0.5ex, -0.5ex) -- ($(\sx ex, -0.5ex)-(0.5ex,0ex)$);
\draw [ffnbclr,  thick] (0.5ex, -1.0ex) -- ($(\sx ex, -1.0ex)-(0.5ex,0ex)$);
\draw [ffnbclr,  thick] (0.5ex, -1.5ex) -- ($(\sx ex, -1.5ex)-(0.5ex,0ex)$);
\draw [ffnbclr,  thick] (0.5ex, -2.0ex) -- ($(\sx ex, -2.0ex)-(0.5ex,0ex)$);
\draw [ffnbclr,  thick] (0.5ex, -2.5ex) -- ($(\sx ex, -2.5ex)-(0.5ex,0ex)$);
\draw [ffnbclr,  thick] (0.5ex, -3.0ex) -- ($(\sx ex, -3.0ex)-(0.5ex,0ex)$);
\draw [ffnbclr,  thick] (0.5ex, -3.5ex) -- ($(\sx ex, -3.5ex)-(0.5ex,0ex)$);
%
%
}
};
\draw [->] (backbone.east) -- (lor.west);
\draw [->] (lor.east) -- ($(lor.east)+(1.40ex,0ex)$) |- (ffnc.west);
\draw [->] (lor.east) -- ($(lor.east)+(1.40ex,0ex)$) |- (ffnb.west);
\draw [->] (ffnc.east) -- ($(mtx1.west)+(2.4ex,-0.0ex)$);
\draw [->] (ffnb.east) -- ($(mtx2.west)+(2.4ex,-0.0ex)$);
}
};
}
};
\node (vslam) [draw=gray, dashed, line width= 0.1pt,dash pattern=on 2pt off 1pt, xshift=-9.2ex, yshift= 0ex,rounded corners=0.3ex, minimum width=13ex, minimum height = 1ex,scale=0.7]{SLAM};
\node (cs) [draw=gray, dashed, line width= 0.1pt,dash pattern=on 2pt off 1pt, xshift=-9.2ex, yshift= -4ex,rounded corners=0.3ex, minimum width=13ex, minimum height = 1ex,scale=0.7]{Control};
\node (vstext) [draw=none, dashed, line width= 0.1pt,dash pattern=on 2pt off 1pt, xshift=1.2ex, yshift= -4.5ex,rounded corners=0.3ex, minimum width=13ex, minimum height = 1ex,scale=0.7]{Perception};
\draw [->] (arm) -| (takeoff);
\draw [->] (takeoff) -| ($(hover.north) + (4.0ex,0ex)$);
\draw [->] ($(hover.south) + (5.8ex,0ex)$) |- (detect.east);
\draw [->] (detect) |- (target);
\draw [->] (target) -| (vs)  node [xshift=13ex,yshift=-2ex]{\footnotesize Yes};
\draw [->] (vs.west) -| ($(grasp.south) - (2ex,0ex)$);
\draw [->] (grasp) |- (delivery);
\draw [->] ($(target.south) - (1ex,0ex)$) |- ($(land.east) + (0ex,0.5ex)$) node [xshift=3ex,yshift=0.9ex]{\footnotesize No};
\draw [->] ($(land.east) - (0ex,0.8ex)$) -- ($(disarm.west) + (0ex,-0.8ex)$);
\draw [<->,dashed,, line width= 0.5pt,dash pattern=on 2pt off 1pt,orange] ($(nwarch.east)-(0.75ex,0ex)$) -| ($(detect.north) - (5.5ex,0.0ex)$);
\draw [->,dashed,, line width= 0.5pt,dash pattern=on 2pt off 1pt,orange] (vslam) -- (cs);
\draw [<->,dashed,, line width= 0.5pt,dash pattern=on 2pt off 1pt,orange] (cs) |- (vs);
\draw [<->,dashed,, line width= 0.5pt,dash pattern=on 2pt off 1pt,orange] (delivery) |- ($(cs.west) + (0ex,0.5ex)$);
\draw [<->,dashed,, line width= 0.5pt,dash pattern=on 2pt off 1pt,orange] (cs.east) -| ($(takeoff.west) - (21.3ex,0.0ex)$) -- (takeoff);
\draw [<->,dashed,, line width= 0.5pt,dash pattern=on 2pt off 1pt,orange] ($(cs.south) - (3ex,0.0ex)$) -- ($(land.north) + (0.75ex,0.0ex)$);
}
};
\end{tikzpicture}
\caption{The proposed state machine}
\label{fig:statemachine}
\vspace{-3ex}
\end{figure}
Despite developing the subsystems, designing a state-machine for the aerial grasping task is complicated due to switching between many phases \cite{loianno2018localization} i.e. detection, visual servoing, grasping, target delivery etc. Hence, we develop a novel end-to-end state machine based on a few novel techniques. It is the final component of our aerial grasping system, which communicates with all the subsystems (Fig.~\ref{fig:infoflow}) to enable the UAV to perform the challenging fruit harvesting task autonomously indoors. The state-machine diagram is shown in Fig.~\ref{fig:statemachine}. Our software is written in C++ and uses Robot Operating System (ROS) for inter-process communication.
\subsection{Hovering}
Precise hovering is critically important, otherwise, it results in camera motion blur, leading to object detection and tracking failure. Our control system achieves precise hovering via thrust microstepping and accelerometer feedback.
\input{figures/grasping_visual_desc}
\subsection{Target Selection and Associated Information}
After achieving stable hovering ($\pm 5$cm), the detector detects fruits in the RGB image acquired from the D$435$i. In our experiments, we choose the target as the fruit instance closest to the UAV in the line of sight. Now we compute the target's $3$D centroid using the D$435$i depth measurements.
\par
\subsection{Target Tracking and Visual Servoing}
The target centroid is used to navigate near the targets, achievable in two ways: (\texttt{i}) open loop, and (\texttt{ii}) closed loop or visual servoing. In the open loop, the target's location is sent only once to the control system, and then navigation is executed in position control mode blindly. For this reason, if UAV drifts due to wind drafts during navigation, the final UAV position may differ from the desired target's location. 
\par
Whereas in the closed loop, the target is continuously tracked in the RGB frames while updating its $3$D centroid. In this case, the control system is set to velocity mode in $x$ (front), while position mode in $y$ (side) and $z$ (vertical). This significantly boosts the system's accuracy. We use color-based tracking to avoid algorithmic complexity. This may fail during overlapped instances but can be easily tackled via depth information and/or deep learning-based tracker, which we leave for future work. 
\subsection{Object Localization via Instance Mapping}
In visual servoing, noisy target centroid is a critical issue. It originates due to noisy depth measurements (Sec.~\ref{sec:intro_inair},~\ref{sec:intro_noisydepth}) and also when UAV can not maintain absolute zero error from the set point. Therefore, we develop \textit{instance mapping} in $3$D, which generates a $3$D map containing the locations of all the instances detected so far. In this strategy, the target location from the object tracker is not directly used; instead, first, the map is updated with the tracker output, and then the location is obtained from the map. Hence, even if the tracker fails during visual servoing, we still have access to the target's location to continue approaching the target. 
\par
\textit{Note:} The UAV localization is independent of the number of apples since it is carried out by our SLAM-based localization. Hence it should not be confused with the apple localization.

\subsection{Grasp Synthesis}
Grasp synthesis refers to generating a grasp pose to grasp a target successfully. In our case, after reaching near the target ($\pm2$cm), the gripper is closed, and the UAV performs a backward motion to separate the fruit from its stem. To execute a grasp sequence, the gripper is opened early, i.e. before the commencement of the visual servoing. The UAV now reaches back to the hovering spot and releases the target. The UAV can also be programmed to deliver the fruit to a prefixed location. However, we only have demonstrated the grasping capabilities. See Fig.~\ref{fig:visdesc} for a visual evolution of the fully autonomous execution of the proposed state machine.
%
%

\section{System Performance}
\label{sec:exp}

We extensively evaluate our \textit{Drone-Bee} system. We perform indoor and outdoor experiments to incorporate the effect of rotor draft, wind turbulence, constrained workspaces, and lighting. The UAV takes off and lands $1.8$m far from the harvesting setup, measured horizontally. Since this is a system paper, we evaluate the overall system performance while discussing only the key results of our subsystems. See \cite{jetsonslam, ffd, tmdc} for a detailed sub-system evaluation.
%
%
\subsection{Localization, Perception and Control}
\label{sec:evalalgo}
Our localization system \cite{jetsonslam} is a quite fast and accurate system. Table~\ref{tab:kaistvio} shows metric accuracy of our method against state-of-the-art systems on the recent KAIST-VIO \cite{kaistvio} dataset for UAVs. Our SLAM system can reach beyond $60$FPS on Jetson-NX at $432\times240$ resolution despite in stereo mode, which is the most unique feature of our localization system. 
\par
Our detector \cite{ffd} outperforms prominent \cite{fasterrcnn} and recent SOTA \cite{yolov8, detr} detectors by having faster training, inference, and promising results on small objects without multi-scale detection (Table.~\ref{tab:detect_perf}, Fig.~\ref{fig:qualitative}). Our detector achieves this performance due to the novel design of its head (Sec.~\ref{sec:detector}). 
\par
\begin{figure}[!t]
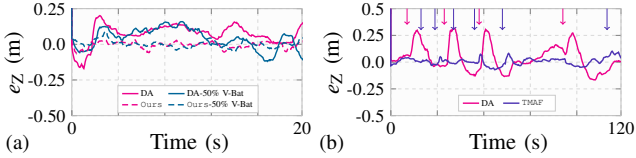

\centering

\colorlet{clr1}{white!0!magenta}
\colorlet{clr2}{blue!60!green}
\colorlet{clr3}{black!40!yellow}
\colorlet{clr4}{blue!60!brown}

\begin{tikzpicture}

\FPeval{\xshfta}{0}
\FPeval{\xshftb}{27}
\FPeval{\xshftc}{0}
\FPeval{\xshftd}{0}

\FPeval{\yshfta}{0-0}
\FPeval{\yshftb}{0-0}
\FPeval{\yshftc}{0-18}
\FPeval{\yshftd}{0-27.9}

\FPeval{\pltw}{53}
\FPeval{\plth}{30}

\FPeval{\scal}{0.45}

\FPeval{\mrksize}{0.3}

\colorlet{clr1}{white!0!magenta}
\colorlet{clr2}{blue!60!green}
\colorlet{clr3}{black!40!yellow}
\colorlet{clr4}{blue!60!brown}

\colorlet{gridclr}{white!85!black}
\colorlet{dlegendclr}{white!80!black}
\colorlet{axisclr}{white!75!black}
\colorlet{axisbgclr}{white!99!black}

\colorlet{dlegendclr}{white!80!black}
\colorlet{legendclr}{white!100!black}
\colorlet{desclr}{white!0!brown}

\FPeval{\lscale}{0.76}
\FPeval{\limscale}{0.8}

\FPeval{\lxshfta}{21.7}
\FPeval{\lxshftb}{21.7}

\FPeval{\lyshfta}{0.5}
\FPeval{\lyshftb}{0.8}

\FPeval{\titlescale}{0.7}
\FPeval{\titlexshift}{21.0}
\FPeval{\titleyshift}{13.7}

\colorlet{gpudclr}{white!80!black}
\colorlet{gpuclr}{white!90!black}
\colorlet{gputxtclr}{white!0!black}

\FPeval{\labelscale}{1.9}
\FPeval{\ticklabelscale}{1.5}

\FPeval{\xlabelxshifta}{0+22}
\FPeval{\xlabelyshifta}{0+1.5}
\FPeval{\ylabelxshifta}{0-3.0}
\FPeval{\ylabelyshifta}{0+9.9}

\FPeval{\xlabelxshiftb}{0+22}
\FPeval{\xlabelyshiftb}{0+1.5}
\FPeval{\ylabelxshiftb}{0-3.0}
\FPeval{\ylabelyshiftb}{0+9.9}

\FPeval{\linew}{1.2}
\FPeval{\dashon}{5.0}
\FPeval{\dashoff}{3.0}

\FPeval{\xshfttakeoff}{0-16}
\FPeval{\xshfthover}{0}
\FPeval{\xshftland}{16}

\node (a) [xshift = \xshfta ex, yshift=\yshfta ex, scale=1.0]{
\scalebox{\scal}
{
\tikz{
\pgfplotsset{width=\pltw ex, height=\plth ex}
\begin{axis}[
   axis background style={fill=axisbgclr},
    title={},
    xlabel={Time (s)},
    ylabel={$e_{\text{Z}}$ (m)},
    xmin=0.0, xmax=0.99,
    ymin=-1.0, ymax=-0.25,
    xtick={0,0.99},
    xticklabels={0,20},
    extra x ticks={0, 0.20, 0.40, 0.60, 0.80, 1.00},
    extra x tick labels={},
    ytick={-1.0, -0.75, -0.50, -0.25},
    yticklabels={-$0.50$, -$0.25$, $0.0$, $0.25$},
     axis line style={axisclr},
    legend image post style={scale =\limscale},
    legend style={at={(\lxshfta ex,\lyshfta ex)},anchor=south, legend columns = 2, draw = {dlegendclr}, fill={legendclr}, nodes={scale=\lscale}},
    legend cell align={left},
    ymajorgrids=true, 
    xmajorgrids=true,
    grid style={dashed, gridclr},
    major tick length=1ex,
    x label style={at={(\xlabelxshifta ex, \xlabelyshifta ex)},scale=\labelscale},
    y label style={at={(\ylabelxshiftb ex, \ylabelyshifta ex)},scale=\labelscale},
    xticklabel style={scale=\ticklabelscale},
    yticklabel style={scale=\ticklabelscale},
    legend cell align={left},
]
\FPeval{\opacty}{0.2}
\input{figures/tmaf_vs_da_comparison/z_da_no_fill}
\input{figures/tmaf_vs_da_comparison/z_da_no_fill_50}
\input{figures/tmaf_vs_da_comparison/z_tmaf_no_fill}
\input{figures/tmaf_vs_da_comparison/z_tmaf_no_fill_50}
\legend{DA~~,DA-$50\%$~V-Bat~~, \texttt{Ours}~~, \texttt{Ours}-$50\%$~V-Bat}
%
\end{axis}
%
}
}};

\node (b) [xshift = \xshftb ex, yshift=\yshftb ex, scale=1.0]{
\scalebox{\scal}
{
\tikz{
\pgfplotsset{width=\pltw ex, height=\plth ex}
\begin{axis}[
   axis background style={fill=axisbgclr},
    title={},
    xlabel={Time (s)},
    ylabel={ $e_{\text{Z}}$ (m)},
    xmin=0.0, xmax=0.99,
    ymin=-1.0, ymax=0.0,
    xtick={0,0.99},
    xticklabels={0,120},
    extra x ticks={0, 0.20, 0.40, 0.60, 0.80, 1.00},
    extra x tick labels={},
    ytick={-1.0, -0.75, -0.50, -0.25, 0.0},
    yticklabels={-$0.5$, -$0.25$, $0.0$, $0.25$, $0.5$},
     axis line style={axisclr},
    legend image post style={scale =\limscale},
    legend style={at={(\lxshftb ex,\lyshftb ex)},anchor=south, legend columns = 4, draw = {dlegendclr}, fill={legendclr}, nodes={scale=\lscale}},
    legend cell align={left},
    ymajorgrids=true, 
    xmajorgrids=true,
    grid style={dashed, gridclr},
    major tick length=1ex,
    x label style={at={(\xlabelxshiftb ex, \xlabelyshiftb ex)},scale=\labelscale},
    y label style={at={(\ylabelxshiftb ex, \ylabelyshiftb ex)},scale=\labelscale},
    xticklabel style={scale=\ticklabelscale},
    yticklabel style={scale=\ticklabelscale},
]
\FPeval{\opacty}{0.2}
\input{figures/disturb_rej__tmaf_vs_da/z_da_no_fill}
\input{figures/disturb_rej__tmaf_vs_da/z_tmaf_no_fill}
\legend{DA~~~, \texttt{TMAF}~~~, $z^\star$}
\draw [->, clr1, thick] (axis cs:0.07,0) -- (axis cs:0.07,-0.15);
\draw [->, clr1, thick] (axis cs:0.23,0) -- (axis cs:0.23,-0.15);
\draw [->, clr1, thick] (axis cs:0.38,0) -- (axis cs:0.38,-0.15);
\draw [->, clr1, thick] (axis cs:0.74,0) -- (axis cs:0.74,-0.15);
\draw [->, clr4, thick] (axis cs:0.13,0) -- (axis cs:0.13,-0.20);
\draw [->, clr4, thick] (axis cs:0.19,0) -- (axis cs:0.19,-0.20);
\draw [->, clr4, thick] (axis cs:0.27,0) -- (axis cs:0.27,-0.20);
\draw [->, clr4, thick] (axis cs:0.48,0) -- (axis cs:0.48,-0.20);
\draw [->, clr4, thick] (axis cs:0.36,0) -- (axis cs:0.36,-0.20);
\draw [->, clr4, thick] (axis cs:0.93,0) -- (axis cs:0.93,-0.20);
\end{axis}
%
}
}};
\node (a) [xshift=-11.7ex, yshift=-5.5ex, scale=0.8]{(a)};
\node (b) [xshift=14.7ex, yshift=-5.5ex, scale=0.8]{(b)};
 
\end{tikzpicture}
\subfloat{\label{fig:tmafvsda}}
\subfloat{\label{fig:tmafvsda_distrub}}
\vspace{-3.7ex}
\newcommand{\varrda}{\raisebox{-0.1ex}{\tikz{\draw[->, clr1] (0,0) -- (0,-0.16);}}}
\newcommand{\varrtmaf}{\raisebox{-0.1ex}{\tikz{\draw[->, clr2] (0,0) -- (0,-0.16);}}}
\caption{Hovering performance of our control system \textit{vs} DA \cite{direct} in different settings. (a) DA achieves an RMSE of $0.16$m but only $0.03$m with TMDC. While for $50\%$ battery discharge, we observe RMSE of $0.15$m with DA and $0.02$m with our control system. (b) Test for external disturbance rejection of $15$N during hovering. The arrows `\protect\varrda' and `\protect\varrtmaf' indicate disturbance introduction. Here DA shows a peak offset of $0.25$m whereas ours only of $0.07$m.}
\label{fig:tmafvsda_hover}
\vspace{-3ex}
\end{figure}

Our control system \cite{tmdc} offers precise hovering, navigation and handles dynamic payloads via accurate thrust estimation. We compare our thrust controller against SOTA thrust controller \cite{direct}, which we outperform. See Figure~\ref{fig:tmafvsda_hover}.
\par
Notably, the individual achievement of each subsystem makes it possible to realize Drone-Bee, which can make its own decisions onboard. Hence, running the entire stack into a limited computing budget is the major novelty and contribution of this paper, which is also listed as a critical bottleneck in \cite{ollero2021past}. By addressing this bottleneck, we push aerial manipulation towards $4^{th}$ generation.
\begin{table}[!t]
\centering
\caption{SLAM evaluation on KAIST-VIO sequences \cite{kaistvio}.}
\label{tab:kaistvio}
\arrayrulecolor{white!70!black}
\scriptsize
\tiny
\setlength{\tabcolsep}{0.6pt}

\vspace{-1ex}
\begin{tabular}{l c c c c c c c c c c c}
\hline 
\multicolumn{1}{c}{\multirow{3}{*}{Approach}} &  \multicolumn{11}{|c}{KAIST-VIO Sequence} \\ \cline{2-12}

 & \multicolumn{3}{|c}{\texttt{circle}}  & \multicolumn{3}{|c}{\texttt{infinite}} & \multicolumn{3}{|c}{\texttt{square}} & \multicolumn{2}{|c}{\texttt{rotation}} \\  \cline{2-12}
 
 & \multicolumn{1}{|c}{\texttt{normal}} & \multicolumn{1}{c}{\texttt{fast}} & \multicolumn{1}{c}{\texttt{head}}  & \multicolumn{1}{|c}{\texttt{normal}} & \multicolumn{1}{c}{\texttt{fast}} & \multicolumn{1}{c}{\texttt{head}}  & \multicolumn{1}{|c}{\texttt{normal}} & \multicolumn{1}{c}{\texttt{fast}} & \multicolumn{1}{c}{\texttt{head}}  & \multicolumn{1}{|c}{\texttt{normal}} & \multicolumn{1}{c}{\texttt{head}} \\  \cline{1-12}
\bdota{}
KIMERA-VIO \cite{kimera}  & \multicolumn{1}{|c}{$0.12$m}   & $0.07$m   & $0.28$m  & \multicolumn{1}{|c}{$0.05$m}   & $0.14$m   & $1.08$m   & \multicolumn{1}{|c}{$0.17$m}   & $0.19$m   & $1.57$m    & \multicolumn{1}{|c}{$0.17$m }   & $0.74$m  \\
\bdota{}
VINS-Fusion + GPU \cite{vinsfusongpu}  & \multicolumn{1}{|c}{$0.09$m}   & $0.13$m   & $0.11$m   & \multicolumn{1}{|c}{$0.09$m}   & $0.05$m   & $0.14$m   & \multicolumn{1}{|c}{$0.12$m}   & $0.11$m   & $0.15$m   & \multicolumn{1}{|c}{$0.12$m}   & $0.11$m \\
\bdota{}
ORB-SLAM$2$ \cite{orb2}  & \multicolumn{1}{|c}{$0.09$m}   & $0.11$m   & $0.13$m   & \multicolumn{1}{|c}{$0.08$m}   & $0.10$m   & $0.12$m   & \multicolumn{1}{|c}{$0.09$m}   & $0.09$m   & $0.16$m   & \multicolumn{1}{|c}{$0.17$m}   & $0.21$m \\

\rowcolor{rwclr}
\bdotb{}
Our Pipeline  & \multicolumn{1}{|c}{$\mathbf{0.014}$m}   & $\mathbf{0.017}$m   & $0.12$m   & \multicolumn{1}{|c}{$\mathbf{0.017}$m}   & $\mathbf{0.016}$m   & $\mathbf{0.09}$m   & \multicolumn{1}{|c}{$\mathbf{0.016}$m}   & $\mathbf{0.017}$m   & $\mathbf{0.04}$m   & \multicolumn{1}{|c}{$\mathbf{0.07}$m}   & $\mathbf{0.09}$m 

\\

\hline
\end{tabular}
\vspace{-1.5ex}
\end{table}
\begin{table}[t]

\centering

\caption{Detection Performance of FFD. AP average precision. `S' and `M' denotes small ($<30\times 30$) and medium-sized ($<90\times 90$ pixels) objects.}
\label{tab:detect_perf}

\arrayrulecolor{white!60!black}
\tiny

\setlength{\tabcolsep}{8.2pt}

\vspace{-0.5ex}
\begin{tabular}{l | c c c c c c c}
\toprule

\multicolumn{1}{c|}{Detector} & AP  &  AP$_{S}$ &  AP$_{M}$ & \makecell{Per Iteration \\ Train-time} & \makecell{Inference \\ time$@$FP$32$} \\ \midrule
 
SSD $@$multi-scale \cite{ssd} & $38.0$  & $20.1$  & $39.1$ & $4.60$s & $32$ms\\ 
DETR $@$single-scale \cite{detr} & $40.2$  & $24.9$ & $43.0$ & $6.80$s  & $25$ms \\ 
Faster-RCNN $@$multi-scale \cite{fasterrcnn} & $45.9$ & $28.7$  & 
$51.5$ & $5.10$s & $49$ms \\ 
YOLO-v$8$ $@$multi-scale \cite{yolov8} & $46.3$ & $29.2$  & 
$51.5$  & $2.65$s  & $29$ms \\ 
YOLO-v$8$ $@$single-scale \cite{yolov8} & $35.2$  & $23.1$ & $41.4$  & $0.95$s  & $24$ms \\ \midrule 
FFD $@$single-scale & $\mathbf{46.6}$ & $\mathbf{31.2}$  & $\mathbf{52.1}$  & $\mathbf{0.40}$s & $\mathbf{11}$ms \\ 

\bottomrule

\end{tabular}
\vspace{-3.0ex}
\end{table}
\input{figures/vision_detections}
\subsection{Evaluation Metrics \& Grasping Performance}
%
We define three error metrics to quantify the grasping performance, as reported in robotic manipulation tasks \cite{acrvsystem2017}.
\subsubsection{Grasp  Success  Rate} A successful grasp occurs when an item is gripped fully or partially by the gripper. A partial grasp occurs when the target does not lie in the grasping cavity of the gripper, i.e. lies outside of the jaw. While full grasping occurs when the object lies completely inside the jaw. To evaluate the grasp success rate, we vary the number of fruits
\par
As shown in Table~\ref{tab:grasp}, Exp-$1$, our system grasps items accurately. A few times, the system was observed to fail a grasp. As per our analysis, depth measurements ($\pm0.03$m) are quite noisy, which resulted in incorrect grasp point. It led to the gripping of the trellis while attempting a grasp. In practice, such kind of trellis is not present except small stems, which do not pose any restriction in front of the system. 
\subsubsection{Average Time per Instance}  It is the average time taken to execute a grasp sequence, measured from target selection to target delivery at a specified spot. To evaluate the average time per instance, we perform two experiments, each with $5$ trails: (\textit{i}) change the number of fruits, and (\textit{ii}), we fix the number of fruits to $5$ but vary their position. Exp-$2$ and Exp-$3$ in Table~\ref{tab:grasp} shows the corresponding results. It can be seen that the average time varies from $\sim 3- 8$ seconds. The smaller time is required when the fruits lie nearly in the line-of-sight, while the larger time is required when the target lie near the boundaries of the harvesting region and the UAV is flying in the middle.
\subsubsection{Error Rate} It is the ratio of a number of grasped items dropped before reaching the delivery spot, and a total number of grasped items. Based on Exp-$1-3$, we did not observe any event which could contribute to the error rate. 
\subsection{Visual Servoing Performance}
We evaluate visual servoing due to its key role in grasping. To setup the experiments, we adhere a fruit target to a stick which can be freely moved. The UAV is instructed to stay $0.20$m before the target while pointing towards it. Now, we move the stick in $x,y,z$ axis simultaneously. We report the deviation of UAV from the desired configuration. We find that the system can perform visual servoing precisely (Table~\ref{tab:vservo}).
\subsection{Frame-Rate \& Computing Resource Occupancy}
\par
Deploying complex autonomy systems onboard is an unsolved challenge \cite{ollero2021past} in transitioning towards $4^{th}$ generation aerial manipulators. Hence, we evaluate our flight and decisional autonomy in terms of computing resources. Table~\ref{tab:timing} shows the maximum frame rates of various autonomy sub-systems and the computer resource profiles, e.g. power consumption and temperature. Notably, despite a tremendous load, $30\%$ of computational space is still left. This clearly caters for the goal of successfully deploying the autonomy engine onboard, indicating the major achievement of this work.
\vspace{-1.0ex}
\subsection{Prototype Hardware Assessment}
We examined gripper durability by continuously opening and closing it for $\sim5$ hours. Although it is a short duration, it is sufficient for a $3$D printed part. Post the experiment, no wear-and-tear was seen in the gripper assembly, indicating that our mechanical designs are stable for proof-of-concept and can be affordably reprinted by researchers interested in this area. We use PLA material for $3$D printing; however, other suitable materials can be chosen for greater strength.
\section{Hidden Challenges, Solutions and Future}
\label{sec:technicalissue}
Apart from the core aerial grasping challenges (Sec.~\ref{sec:intro}), we also introduce the hidden challenges faced during real-time implementation. We also discuss the prospects of the paper.
\subsubsection{A Coaxial Rotor UAV}
In our UAV, the long arm introduces off-center load intensively. The long arm was required to maintain a safety gap of $30$cm between the rotors and the gripper, given the large size of the UAV used. If not accounted for, the propellers may hit nearby structures while executing a grasp sequence. Thus a short arm is more beneficial, but it requires the UAV to span a smaller area. This issue can be resolved via a coaxial rotor UAV.
\subsubsection{Gripper Improvements}
Although the gripper developed is sufficient for this task, we observed that if the fruit instance is attached to the trellis firmly, the UAV needs to exert a pull effort to detach the target. This issue can be resolved by improvising the gripper with a rotatable wrist.
%
%
%

%
%
%
%
%
%
%
%
%
%
%
%
%
%
%
%
%
%
%
%
%

\begin{table}[t]

\centering
\caption{Grasping Performance Evaluation.}
\label{tab:grasp}

\arrayrulecolor{white!60!black}
\tiny

\setlength{\tabcolsep}{8.2pt}

\vspace{-0.5ex}
\begin{tabular}{c | c  c  c  c c }
\hline

\multicolumn{2}{c}{Exp-$1$} & \multicolumn{2}{||c}{Exp-$2$  $@$ Variable \#Fruits} & \multicolumn{2}{||c}{Exp-$3$ $@$ Variable Position} \\ \hline

\#Fruits & \makecell{Grasp \\ Success rate ($\%$)}  & \multicolumn{1}{||c|}{\#Fruits} & \makecell{Average \\ Grasping Time}   & \multicolumn{1}{||c|}{\#Fruits} & \makecell{Average \\ Grasping Time}  \\ \hline

$2$ & $100\%$ & \multicolumn{1}{||c|}{$2$} & $4$s  & \multicolumn{1}{||c|}{$5$} & $9$s \\ 
$3$ & $66\%$ & \multicolumn{1}{||c|}{$3$} & $6$s  & \multicolumn{1}{||c|}{$5$} & $11$s   \\ 
$6$ & $80\%$ & \multicolumn{1}{||c|}{$6$} & $5$s  & \multicolumn{1}{||c|}{$5$} & $8$s   \\ 
$8$ & $87\%$ & \multicolumn{1}{||c|}{$8$} & $4$s  & \multicolumn{1}{||c|}{$5$} & $7$s   \\ \hline

\end{tabular}
\vspace{-3ex}
\end{table}
\begin{table}[t]

\centering
\caption{Visual servoing performance.}
\label{tab:vservo}

\arrayrulecolor{white!60!black}
\tiny

\setlength{\tabcolsep}{14.5pt}
\vspace{-0.5ex}

\begin{tabular}{c | c | c | c | c}
\hline

Metric  & $\Delta x$ & $\Delta y$ & $\Delta z$ & $\Delta \psi$  \\ \hline

mean ($\mu$) & $0.021$m  & $0.029$m  & $0.018$m  & $2^{o}$  \\ 
standard-deviation ($\sigma$) & $0.019$m  & $0.023$m  & $0.020$m  & $3^{o}$  \\ 

\hline

\end{tabular}
\vspace{-1.0ex}
\end{table}
\begin{table}[!t]

\centering
\caption{Frame processing rates and computing resource utilization.}
\label{tab:timing}

\arrayrulecolor{white!60!black}
\tiny

\setlength{\tabcolsep}{11.2pt}
\vspace{-0.5ex}

\begin{tabular}{c | c || c | c}
\hline

Algorithmic Component & Frame rate  & \makecell{Computing \\Resource} & Utilization  \\ \hline

Stereo Image Acquisition & $30$ FPS $@432\times240$ & RAM & $40\%$ \\ 
Image Rectification & $1000$ FPS $@432\times240$ & CPU & $70\%$ \\ 
SLAM & $60$ FPS $@432\times240$ & GPU & $45\%$ \\ 
Detector & $100$ FPS $@320\times240$ & Power & $20$W \\ 
Realsense Image Acquisition & $60$ FPS $@320\times240$ & Temperature & $55^{o}$C \\ 

\hline

\end{tabular}
\vspace{-4.5ex}
\end{table}
\subsubsection{Depth Sensing}
One major cause of grasp failures was inaccurate depth sensing. Intel D$435$i provides noisy depth, and due to which, even hovering at the exact location, the depth of the target instance is measured $\sim 2-3$cm in or out of the harvesting region, leading to the gripping of trellis woods. A potential solution can be to use stereo triangulation.
\subsubsection{Hardware Synchronized Stereo-Rig}
We strongly recommend a hardware synchronized stereo-rig where both of the cameras are triggered at the same time. It is so because, sometimes, due to the operating system overhead or consumption of computing resources, it is not possible to grab images from each camera in a time-synchronized manner. Hence, two images may have different capture-time which negatively affects rectification and stereo matching, thus causing errors in the $6$DoF pose estimation.
%
%
\subsubsection{Depth Measurements Rate}
We used registered depth which is aligned w.r.t. to the RGB camera. By default, the registered depth is computed on the host CPU instead of the ASIC of D$435$i, which can provide the depth images at a rate of only $\sim 2-3$ FPS on Jetson-NX. We solved this issue by compiling Intel Realsense SDK with CUDA to enable GPU operations. By doing so, we achieve a rate of $\sim30$ FPS. 
\section{Conclusion}
\label{sec:conc}
This work introduces comprehensive hardware design and flight autonomy engine for off-center aerial grasping in GPS-denied environments. The contributions of the paper are: to present aerial harvesting challenges, intricate details of the hardware designs and in-lab fabrication, sensor selection and integration, experimental setup realization, and system integration along with developing crucial sub-systems such as object detection, positioning system, and control system. The flight autonomy engine can run on a $\sim10$W NVIDIA Jetson-NX embedded computer which is the representative achievement of this paper to push aerial manipulators state-of-the-art towards $4^{th}$ generation. We call the resulting system as \system{}, which is demonstrated for a challenging task of fruit harvesting. We evaluate \system{} by conducting several experiments and show that it is capable of grasping desired items precisely. With its capabilities, the \system{} system opens new doors to the future of autonomous aerial agriculture.

%
\bibliographystyle{ieeetr}
\scriptsize
\bibliography{bibfile}

\begin{thebibliography}{10}

\bibitem{towards}
A.~Kumar, M.~Vohra, R.~Prakash, and L.~Behera, ``Towards deep learning assisted
  autonomous {UAV}s for manipulation tasks in gps-denied environments,'' in
  {\em 2020 IEEE/RSJ International Conference on Intelligent Robots and Systems
  (IROS)}, pp.~1613--1620, IEEE, 2020.

\bibitem{semi}
A.~Kumar and L.~Behera, ``Semi supervised deep quick instance detection and
  segmentation,'' in {\em 2019 International Conference on Robotics and
  Automation (ICRA)}, pp.~8325--8331, IEEE, 2019.

\bibitem{acrvsystem2017}
D.~Morrison, A.~W. Tow, M.~McTaggart, R.~Smith, N.~Kelly{-}Boxall,
  S.~Wade{-}McCue, J.~Erskine, R.~Grinover, A.~Gurman, T.~Hunn, D.~Lee,
  A.~Milan, T.~Pham, G.~Rallos, A.~Razjigaev, T.~Rowntree, K.~Vijay, Z.~Zhuang,
  C.~F. Lehnert, I.~D. Reid, P.~Corke, and J.~Leitner, ``Cartman: The low-cost
  cartesian manipulator that won the amazon robotics challenge,'' 2017.

\bibitem{ollero2021past}
A.~Ollero, M.~Tognon, A.~Suarez, D.~Lee, and A.~Franchi, ``Past, present, and
  future of aerial robotic manipulators,'' {\em IEEE Transactions on Robotics},
  vol.~38, no.~1, pp.~626--645, 2021.

\bibitem{nimbrombzirc2020}
M.~Beul, M.~Schwarz, J.~Quenzel, M.~Splietker, S.~Bultmann, D.~Schleich,
  A.~Rochow, D.~Pavlichenko, R.~A. Rosu, P.~Lowin, {\em et~al.}, ``Target
  chase, wall building, and fire fighting: Autonomous {UAV}s of team nimbro at
  {MBZIRC} 2020,'' {\em arXiv preprint arXiv:2201.03844}, 2022.

\bibitem{lee2021closed}
L.~Y. Lee, O.~A. Syadiqeen, C.~P. Tan, and S.~G. Nurzaman, ``Closed-structure
  compliant gripper with morphologically optimized multi-material fingertips
  for aerial grasping,'' {\em IEEE Robotics and Automation Letters}, vol.~6,
  no.~2, pp.~887--894, 2021.

\bibitem{mclaren2019passive}
A.~McLaren, Z.~Fitzgerald, G.~Gao, and M.~Liarokapis, ``A passive closing,
  tendon driven, adaptive robot hand for ultra-fast, aerial grasping and
  perching,'' in {\em 2019 IEEE/RSJ International Conference on Intelligent
  Robots and Systems (IROS)}, pp.~5602--5607, IEEE, 2019.

\bibitem{ruggiero2015multilayer}
F.~Ruggiero, M.~A. Trujillo, R.~Cano, H.~Ascorbe, A.~Viguria, C.~Per{\'e}z,
  V.~Lippiello, A.~Ollero, and B.~Siciliano, ``A multilayer control for
  multirotor {UAV}s equipped with a servo robot arm,'' in {\em International
  conf. on robotics and automation (ICRA)}, pp.~4014--4020, IEEE, 2015.

\bibitem{fasterrcnn}
S.~Ren, K.~He, R.~Girshick, and J.~Sun, ``{Faster R-CNN}: Towards real-time
  object detection with region proposal networks,'' in {\em Advances in neural
  information processing systems}, pp.~91--99, 2015.

\bibitem{ssd}
W.~Liu, D.~Anguelov, D.~Erhan, C.~Szegedy, S.~Reed, C.-Y. Fu, and A.~C. Berg,
  ``{SSD}: Single shot multibox detector,'' in {\em European conference on
  computer vision}, pp.~21--37, Springer, 2016.

\bibitem{fpn}
T.-Y. Lin, P.~Doll{\'a}r, R.~Girshick, K.~He, B.~Hariharan, and S.~Belongie,
  ``Feature pyramid networks for object detection,'' {\em CVPR}, 2017.

\bibitem{detr}
N.~Carion, F.~Massa, G.~Synnaeve, N.~Usunier, A.~Kirillov, and S.~Zagoruyko,
  ``End-to-end object detection with transformers,'' in {\em European Conf. on
  Computer Vision}, pp.~213--229, Springer, 2020.

\bibitem{yolov8}
``{YOLO-v8},'' in {\em https://github.com/ultralytics/ultralytics}.

\bibitem{orb2}
R.~Mur-Artal and J.~D. Tard{\'o}s, ``{ORB-SLAM2}: An open-source slam system
  for monocular, stereo, and rgb-d cameras,'' {\em IEEE Transactions on
  Robotics}, vol.~33, no.~5, pp.~1255--1262, 2017.

\bibitem{dmvio}
L.~Von~Stumberg and D.~Cremers, ``{DM-VIO}: Delayed marginalization
  visual-inertial odometry,'' {\em IEEE Robotics and Automation Letters}, 2022.

\bibitem{pounds2012stability}
P.~E. Pounds, D.~R. Bersak, and A.~M. Dollar, ``Stability of small-scale {UAV}
  helicopters and quadrotors with added payload mass under pid control,'' {\em
  Autonomous Robots}, vol.~33, no.~1, pp.~129--142, 2012.

\bibitem{direct}
M.~Hamandi, M.~Tognon, and A.~Franchi, ``Direct acceleration feedback control
  of quadrotor aerial vehicles,'' in {\em 2020 IEEE International Conference on
  Robotics and Automation (ICRA)}, pp.~5335--5341, IEEE.

\bibitem{heredia2014control}
G.~Heredia, A.~Jimenez-Cano, I.~Sanchez, D.~Llorente, V.~Vega, J.~Braga,
  J.~Acosta, and A.~Ollero, ``Control of a multirotor outdoor aerial
  manipulator,'' in {\em 2014 IEEE/RSJ international conference on intelligent
  robots and systems}, pp.~3417--3422, IEEE, 2014.

\bibitem{suarez2017anthropomorphic}
A.~Suarez, P.~R. Soria, G.~Heredia, B.~C. Arrue, and A.~Ollero,
  ``Anthropomorphic, compliant and lightweight dual arm system for aerial
  manipulation,'' in {\em 2017 IEEE/RSJ International Conference on Intelligent
  Robots and Systems (IROS)}, pp.~992--997, IEEE, 2017.

\bibitem{stephens2022aerial}
B.~Stephens, L.~Orr, B.~B. Kocer, H.-N. Nguyen, and M.~Kovac, ``An aerial
  parallel manipulator with shared compliance,'' {\em IEEE Robotics and
  Automation Letters}, vol.~7, no.~4, pp.~11902--11909, 2022.

\bibitem{corsini2022nonlinear}
G.~Corsini, M.~Jacquet, H.~Das, A.~Afifi, D.~Sidobre, and A.~Franchi,
  ``Nonlinear model predictive control for human-robot handover with
  application to the aerial case,'' in {\em IEEE/RSJ International Conference
  on Intelligent Robots and Systems (IROS)}, pp.~7597--7604, IEEE, 2022.

\bibitem{loianno2018localization}
G.~Loianno, V.~Spurny, J.~Thomas, T.~Baca, D.~Thakur, D.~Hert, R.~Penicka,
  T.~Krajnik, A.~Zhou, A.~Cho, {\em et~al.}, ``Localization, grasping, and
  transportation of magnetic objects by a team of mavs in challenging
  desert-like environments,'' {\em IEEE Robotics and Automation Letters},
  vol.~3, no.~3, pp.~1576--1583, 2018.

\bibitem{minneapple}
N.~H{\"a}ni, P.~Roy, and V.~Isler, ``Minneapple: a benchmark dataset for apple
  detection and segmentation,'' {\em IEEE Robotics and Automation Letters},
  vol.~5, no.~2, pp.~852--858, 2020.

\bibitem{deepfruit}
S.~Bargoti and J.~Underwood, ``Deep fruit detection in orchards,'' in {\em 2017
  IEEE international conference on robotics and automation (ICRA)},
  pp.~3626--3633, IEEE, 2017.

\bibitem{surveying}
P.~Roy and V.~Isler, ``Surveying apple orchards with a monocular vision
  system,'' in {\em 2016 IEEE international conference on automation science
  and engineering (CASE)}, pp.~916--921, IEEE, 2016.

\bibitem{opencv}
G.~Bradski and A.~Kaehler, {\em Learning OpenCV: Computer vision with the
  OpenCV library}.
\newblock " O'Reilly Media, Inc.", 2008.

\bibitem{jetsonslam}
A.~Kumar, J.~Park, and L.~Behera, ``High-speed stereo visual slam for
  low-powered computing devices,'' {\em IEEE Robotics and Automation Letters},
  2023.

\bibitem{tmdc}
A.~Kumar and L.~Behera, ``Thrust microstepping via acceleration feedback in
  quadrotor control for aerial grasping of dynamic payload,'' {\em IEEE
  Robotics and Automation Letters}, 2023.

\bibitem{geometrictracking}
T.~Lee, M.~Leok, and N.~H. McClamroch, ``Geometric tracking control of a
  quadrotor {UAV} on se (3),'' in {\em 49th IEEE conference on decision and
  control (CDC)}, pp.~5420--5425, IEEE, 2010.

\bibitem{ffd}
A.~Kumar and L.~Behera, ``High-speed detector for low-powered devices in aerial
  grasping,'' {\em IEEE Robotics and Automation Letters}, 2024.

\bibitem{vgg}
K.~Simonyan and A.~Zisserman, ``Very deep convolutional networks for
  large-scale image recognition,'' {\em CoRR}, vol.~abs/1409.1556, 2014.

\bibitem{kaistvio}
J.~Jeon, S.~Jung, E.~Lee, D.~Choi, and H.~Myung, ``Run your visual-inertial
  odometry on nvidia jetson: Benchmark tests on a micro aerial vehicle,'' {\em
  IEEE Robotics and Automation Letters}, vol.~6, no.~3, pp.~5332--5339, 2021.

\bibitem{kimera}
A.~Rosinol, M.~Abate, Y.~Chang, and L.~Carlone, ``Kimera: an open-source
  library for real-time metric-semantic localization and mapping,'' in {\em
  2020 IEEE International Conference on Robotics and Automation (ICRA)},
  pp.~1689--1696, IEEE, 2020.

\bibitem{vinsfusongpu}
``Vins-fusion-gpu,'' in {\em https://github.com/pjrambo/VINS-Fusion-gpu}.

\end{thebibliography}

\end{document}